\documentclass[runningheads]{llncs}

\usepackage{eccv}

\usepackage{eccvabbrv}

\usepackage{graphicx}
\usepackage{booktabs}

\usepackage[accsupp]{axessibility}  

\usepackage{multirow}
\usepackage{soul}
\usepackage[normalem]{ulem}
\usepackage{wrapfig}
\usepackage[misc]{ifsym} 

\usepackage[pagebackref,breaklinks,colorlinks,citecolor=eccvblue]{hyperref}

\usepackage{orcidlink}

\begin{document}
	\title{GPSFormer: A Global Perception and Local Structure Fitting-based Transformer for Point Cloud Understanding} 
	
	\titlerunning{GPSFormer}
	
	
	\author{Changshuo Wang\inst{1}\orcidlink{0000-0002-4056-4922} \and
	Meiqing Wu\inst{1} \and
	Siew-Kei Lam\inst{1,2}$^{\textrm{\Letter}}$\orcidlink{0000-0002-8346-2635} \and
	Xin Ning\inst{3}\orcidlink{0000-0001-7897-1673} \and
	Shangshu Yu\inst{1}\orcidlink{0000-0002-5000-0979} \and
	Ruiping Wang\inst{1}\orcidlink{0000-0002-9576-8164} \and
	Weijun Li\inst{3}\orcidlink{0000-0001-9668-2883} \and
	Thambipillai Srikanthan\inst{1,2}}

\authorrunning{C.Wang et al.}


\institute{
	Cyber Security Research Center (CYSREN), Nanyang Technological University \\
	\and
	College of Computing and Data Science, Nanyang Technological University \\
	\and
	Institute of Semiconductors, Chinese Academy of Sciences \\
	\email{\{changshuo.wang, meiqingwu, assklam, shangshu.yu, ruiping.wang, astsrikan\}@ntu.edu.sg}, 
	\email{\{ningxin, wjli\}@semi.ac.cn}
}

\maketitle
	
	\let\thefootnote\relax\footnotetext{$^{\textrm{\Letter}}$ The corresponding author.}

	\begin{abstract}
		Despite the significant advancements in pre-training methods for point cloud understanding, directly capturing intricate shape information from irregular point clouds without reliance on external data remains a formidable challenge. To address this problem, we propose \textbf{GPSFormer}, an innovative \textbf{G}lobal \textbf{P}erception and Local \textbf{S}tructure \textbf{F}itting-based Transf\textbf{ormer}, which learns detailed shape information from point clouds with remarkable precision. The core of GPSFormer is the Global Perception Module (GPM) and the Local Structure Fitting Convolution (LSFConv). Specifically, GPM utilizes Adaptive Deformable Graph Convolution (ADGConv) to identify short-range dependencies among similar features in the feature space and employs Multi-Head Attention (MHA) to learn long-range dependencies across all positions within the feature space, ultimately enabling flexible learning of contextual representations. Inspired by Taylor series, we design LSFConv, which learns both low-order fundamental and high-order refinement information from explicitly encoded local geometric structures. Integrating the GPM and LSFConv as fundamental components, we construct GPSFormer, a cutting-edge Transformer that effectively captures global and local structures of point clouds. Extensive experiments validate GPSFormer's effectiveness in three point cloud tasks: shape classification, part segmentation, and few-shot learning. The code of GPSFormer is available at \url{https://github.com/changshuowang/GPSFormer}.
		
	\end{abstract}

	\section{Introduction}
	
	In recent years, point cloud understanding~\cite{sun2020scalability} techniques have been widely applied in fields such as autonomous driving\cite{wang2022learning, wangchangshuo20223d}, robotics\cite{fang2023you, zhang2023deep}, and public safety\cite{wang20233d, fang2022multi}. However, due to the unordered and irregular nature of point clouds, effectively extracting inherent shape information from them remains an extremely challenging research topic. Accurately and efficiently learning shape perception from point clouds has emerged as a prominent and noteworthy problem.
	
	Early researches~\cite{zhu2013facade,li2020end,wei2020view,yang2019learning, maturana2015voxnet, riegler2017octnet} converted point cloud data into multi-view\cite{yu2024pedestrian, fang2021unbalanced, fang2020v, fang2021animc} or voxel representations, utilizing traditional convolutional neural networks to learn shape information. However, this conversion process often led to the loss 
	\begin{wrapfigure}{r}{0.5\textwidth}
	\centering
	\includegraphics[width=1.0\linewidth]{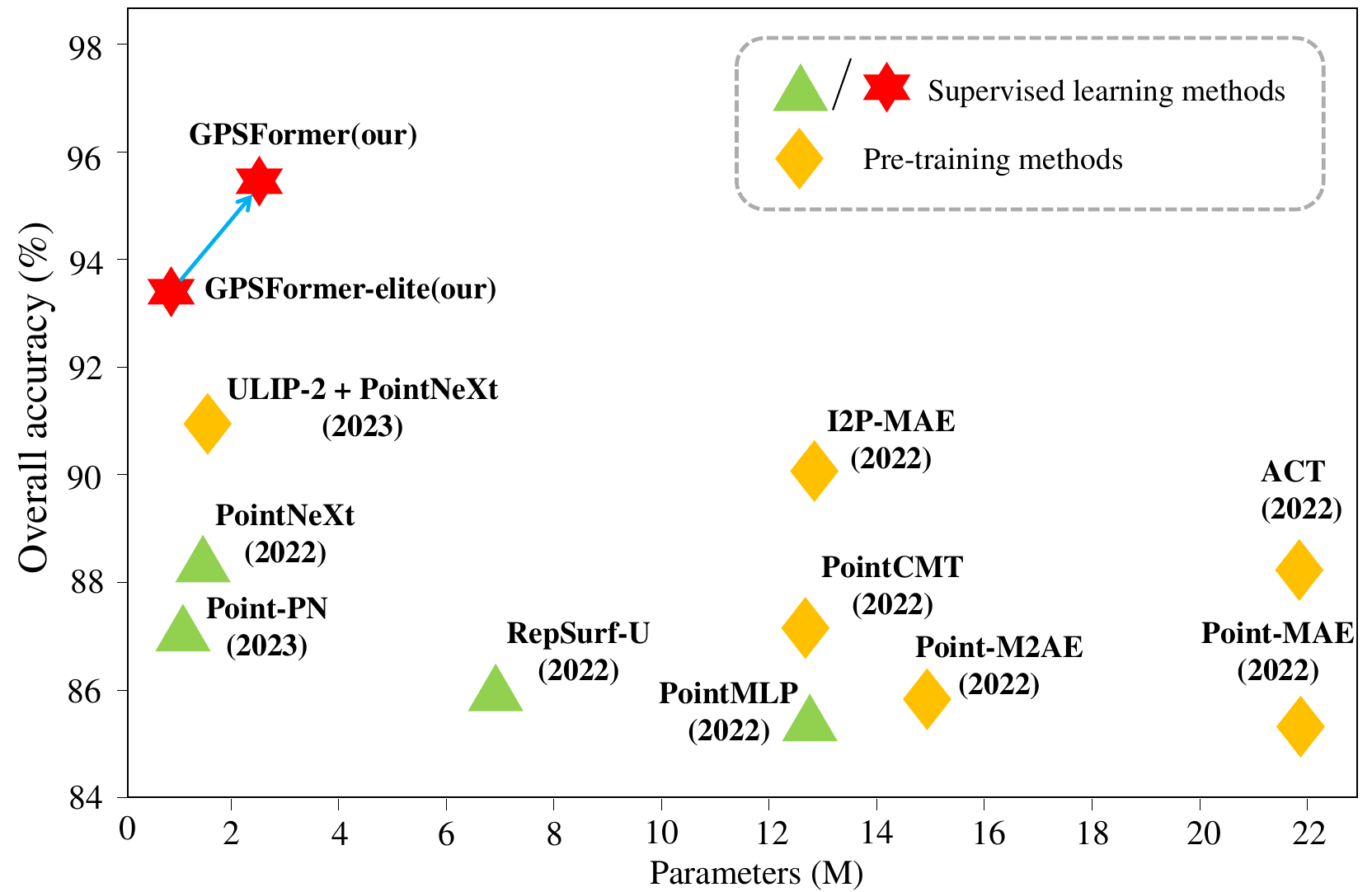}
	\caption{Performance comparison on the challenging ScanobjectNN dataset. We show supervised learning-based and pre-training-based methods with parameters less than 22M. The proposed supervised learning GPSFormer outperforms state-of-the-art methods, achieving an accuracy of 95.4\% with a modest parameter of 2.36M.}
	\label{f1}
	\end{wrapfigure}
	of inherent geometric information and incurred high computational costs. PointNet~\cite{qi2017pointnet} directly encoded each point in the point cloud independently and aggregated global features through max pooling, but this approach overlooked local structural information. To address this issue, subsequent works~\cite{wang2019dynamic,wang2018local,komarichev2019cnn,rao2020global} proposed a series of methods based on local feature aggregation. These methods divide the point cloud into different local subsets through farthest point sampling (FPS), then learn local shape representations by constructing local aggregation operators, and finally learn from local to global shape perception by constructing a hierarchical structure. However, such methods overlook the long-range dependency relationships among points.
	
	Some researchers~\cite{zhang2021pvt, engel2021point, guo2021pct,zhao2021point} have utilized the powerful long-range dependency learning capabilities of Transformer\cite{fang2023hierarchical, fang2023annotations} and applied this structure to point cloud analysis. For example, Point Transformer~\cite{zhao2021point} uses self-attention in local neighborhoods to learn long-range dependencies among points. PCT~\cite{guo2021pct} proposes an offset attention module to learn global context representations. Nevertheless, Transformers that consider both short-range dependencies and long-range dependency relationships, as well as local structure modeling, have rarely been explored. With the rapid development of self-supervised learning and large language models\cite{kasneci2023chatgpt}, some researchers~\cite{yu2022point,pang2022masked,xue2023ulip,ren2016recon,chen2023pointgpt} have proposed a series of methods based on pre-training or multimodal large language models\cite{fang2024fewer}. Although these approaches have improved performance by utilizing external data to assist point cloud models, they have not completely solved the problem of point cloud structural representation.
	
	To overcome the limitations, we propose GPSFormer to learn rich contextual shape perception from point clouds. GPSFormer consists of two core components: a Global Perception Module (GPM) and a Local Structure Fitting Convolution (LSFConv). Within the GPM, we introduce the Adaptive Deformable Graph Convolution (ADGConv) which empowers point features to dynamically navigate the entirety of the point cloud feature space. This allows for the flexible construction of suitable local neighborhoods, facilitating the learning of strong feature representations and short-range dependencies for similar structures. Following this, the features, both pre- and post-transformation, are fed into a Residual Cross-Attention (RCA), enriching the context structural understanding. Conclusively, the model harnesses a multi-head attention (MHA) to capture the long-range dependencies inherent in point clouds. Inspired by Taylor series, we design the LSFConv which treats local structure representation as a polynomial fitting problem to precisely capture subtle changes in local geometric information. Specifically, low-order terms are employed to fit the flat parts of the local structure, typically encompassing the basic shapes and overall trends of the point cloud. High-order terms are used to fit the edges and detailed parts of the local structure, thus capturing complex variations and fine features. As shown in \cref{f1}, GPSFormer achieves excellent performance.
	
	The main contributions of this paper are as follows:
	\begin{itemize}
		\item We propose GPSFormer, a global perception and local structure fitting-based transformer, to learn rich contextual information and precise shape perception from irregular point clouds.
		
		\item We introduce the novel GPM and LSFConv. GPM learns both short-range and long-range point dependencies, while the Taylor series-inspired LSFConv captures low and high-frequency local geometric information through polynomial fitting.

		\item The proposed GPSFormer achieves state-of-the-art results in three point cloud tasks, notably exceeding the current best supervised learning method by 5.0\% in accuracy on the challenging ScanObjectNN dataset.
	\end{itemize}
	
	\section{Related Works}
	
	\subsection{Indirect-based Representation Methods.} Early methods~\cite{zhu2013facade,li2020end,wei2020view,yang2019learning, maturana2015voxnet, riegler2017octnet} transformed unstructured point clouds into multi-views or voxels for 3D shape learning. Multi-view-based methods~\cite{zhu2013facade,li2020end,wei2020view,yang2019learning} converted point clouds into 2D multi-view images, using 2D convolutional neural networks (CNNs). For instance, MVCNN~\cite{su2015multi} integrated information from various perspectives into global features describing 3D objects. However, these methods faced challenges in viewpoint selection and fusion, leading to inherent loss of geometric structure information. Voxel-based methods~\cite{maturana2015voxnet, riegler2017octnet} used 3D convolution to extract shape information from voxel grids. However, voxelization incurred significant computational overhead, and the resolution of voxels resulted in the loss of 3D shape information.
	
	\subsection{Direct-based Representation Methods.} To address these issues, PointNet~\cite{qi2017pointnet} pioneered direct deep learning for point clouds. However, it overlooked local structure information. To address this issue, PointNet++~\cite{qi2017pointnet++} grouped point clouds into different local neighborhoods through Furthest Point Sampling (FPS) and performed feature aggregation within each local neighborhood. Within the "Sampling-Grouping-Aggregation" framework, existing works~\cite{wang2019dynamic,wang2018local,komarichev2019cnn,rao2020global} have designed methods based on point-wise Multi-Layer Perceptron (MLP), convolution operations, and attention mechanisms. Some approaches~\cite{liu2019relation, wu2019pointconv, thomas2019kpconv, jiang2023lttpoint, zhang2024pointgt} enhanced model potential by designing local feature encoding and network structures. Recent works~\cite{zhang2021pvt, engel2021point, guo2021pct,zhao2021point,park2023self} significantly improved point cloud understanding by designing different Transformer structures to learn long-range dependencies in point clouds. However, challenges persisted in simultaneously capturing global context and local information.
	
	In contrast to the above methods, our proposed Adaptive Deformable Graph Convolution (ADGConv) flexibly learns short-range dependencies among similar features. The introduced Local Structure Fitting Convolution (LSFConv) employs a Taylor series fitting approach to finely analyze the local structures and details of point clouds.
	
	\subsection{Pre-training-based Representation Methods.}
	Recent research~\cite{yu2022point,pang2022masked,xue2023ulip,ren2016recon,chen2023pointgpt} has attempted to leverage multimodal data, such as point cloud, text and images, to pre-train point cloud models for downstream tasks, significantly enhancing shape perception capabilities. For example, ULIP~\cite{xue2023ulip} used multimodal information like images, text, and 3D point clouds to learn a unified representation space for objects or scenes, improving 3D point cloud understanding. PointGPT~\cite{chen2023pointgpt}, an auto-regressive generation method, partitioned point clouds into blocks and used a Transformer-based decoder and dual masking strategy to learn latent representations, predicting the next point to address point cloud-related challenges.
	
	Although the use of multimodal pre-training effectively enhanced downstream point cloud understanding tasks, it has a high demand for data and a long training time. Furthermore, the upper limit of the pre-training effect is still constrained by the expressive power of the point cloud model structure.
	
	\section{Method}
	
	As shown in \cref{f3}, we provide a detailed exposition of the proposed GPSFormer. This section is structured as follows: Firstly, we review point convolution (~\cref{31}). Secondly, we introduce the Global Perception Module (GPM) (~\cref{32}). Thirdly, we propose the Local Structure Fitting Convolution (LSFConv) (~\cref{33}). Fourthly, based on GPM and LSFConv, we introduce the GPSFormer architecture and its application details (~\cref{34}).
	
	\subsection{Background}
	\label{31}
	
	Current methods based on local feature aggregation typically follow a structure design of "sampling-grouping-aggregation" to construct local feature extraction blocks. The sampling operation utilizes the Farthest Point Sampling (FPS) method to downsample the input point cloud to create a representative set of sampled points. The grouping operation often employs K-Nearest Neighbors (KNN) or spherical queries to build local neighborhoods for each representative sampled point. The aggregation step uses mapping functions and max pooling to obtain local shape representations.
	
	For each stage, we assume that representative sampled points can be described as \(\{(p_i, f_i)\}_{i=1}^M\), where $M$ represents the number of points,  \(p_i \in \mathbb{R}^{1 \times 3}\) and \(f_i \in \mathbb{R}^{1 \times C}\) represent the coordinates and features of the $i$-th point, respectively. $C$ denotes the number of feature channels. The point convolution for local feature aggregation can be formalized as:
	
	\begin{equation}
		f_i^{\prime}=\mathcal{A}\left(\left\{\mathcal{M}\left(p_i, p_j\right) \cdot \mathcal{T}\left(f_i, f_j\right) \mid p_j \in \mathcal{N}\left(p_i\right)\right\}\right),
	\end{equation}
	where $\mathcal{A}$ is the aggregation function, usually max pooling. $\mathcal{M}$ and $\mathcal{T}$ are mapping functions, typically implemented as Multi-Layer Perceptrons (MLPs). \(N(p_i)\) represents the local neighborhood of \(p_i\), and \(p_j\) are neighboring points of \(p_i\).

	Another aggregation method is to aggregate in the feature space, commonly known as dynamic graph convolution (DGC)~\cite{wang2019dynamic}. It aggregates local features in the feature space, meaning that points close in the feature space may be far apart in coordinate space. DGC can be formalized as:
	\begin{equation}
		f_i^{\prime}=\mathcal{A}\left(\left\{\mathcal{T}\left(f_i, f_j\right) \mid f_j \in \mathcal{N}\left(f_i\right)\right\}\right).
	\end{equation}

	\subsection{Global Perception Module}
	\label{32}
	
	Directly aggregating local features of point clouds may struggle to capture meaningful shape information. We find that modeling the global context of point features before local feature aggregation helps to obtain robust shape perception. 
	
	Therefore, we have developed a Global Perception Module (GPM), which initially employs the innovative Adaptive Deformable Graph Convolution (ADGConv) to reinforce short-range dependencies among similar features within the feature space. Subsequently, it utilizes Residual Cross-Attention (RCA) and Multi-Head Attention (MHA) to capture long-range dependencies across all positions within the feature space. GPM provides guidance for subsequent local structure fitting.
	
	Firstly, dynamic graphs are typically constructed using KNN in the feature space, which makes the receptive field of the dynamic graph easily influenced by the number of neighboring points \(K\). When \(K\) is small, the receptive field of the dynamic graph focuses on the local coordinate neighborhood. When \(K\) is large, the receptive field of the dynamic graph is distributed over some semantically unrelated points, making it challenging to learn distinguishable feature representations within similar components.
	
	To tackle this issue, we introduce the ADGConv. Initially, we define a feature offset \(\Delta(f_i)\) for sampled points, allowing them to traverse the entire feature space and flexibly construct appropriate local neighborhoods. The offset \(\Delta(f_i)\) is adaptively acquired for representative sampling points \(f_i\) through a learnable feature transformation function \(\phi\), indicating the preference of \(f_i\) for a specific position in the feature space. The feature \(\hat{f}_i\) after transformation can be obtained by \(f_i\) and \(\Delta(f_i)\). And we use \(\hat{f}_i\) as the central point, with the original feature \(f_i\) defining the sampling space for constructing local neighborhoods, to build a dynamic graph for obtaining the enhanced feature \(f_i^d\). This roaming process of ADGConv aids in learning robust feature representations among similar components. The ADGConv can be formalized as follows:
	
	\begin{equation}
		\Delta\left(f_i\right)=\phi\left(f_i\right)
	\end{equation}
	
	\begin{equation}
		\hat{f}_i=f_i+\Delta\left(f_i\right)
	\end{equation}
	
	\begin{equation}
		\mathcal{T}\left(\left(\hat{f}_i\right), f_j\right)=\psi\left(\left[\hat{f}_i, f_j-\hat{f}_i\right]\right)
	\end{equation}
	
	\begin{equation}
		f_i^a=\mathcal{A}\left(\left\{\mathcal{T}\left(\widehat{f}_i, f_j\right) \mid f_j \in \mathcal{N}\left(\widehat{f}_i\right)\right\}\right),
	\end{equation}
	where \(\phi\) and \(\psi\) denote MLPs, and $[\cdot]$ represents concatenation. \(f_i^a \in \mathbb{R}^{1 \times C}\) is the output of ADGConv. 
	
	Next, the RCA fuses the displaced feature \( \hat{f}_i \) and the output feature \( f_i^a \) of ADGConv by cross-attention, with the formula 
	\begin{equation}
		f_i^r = f_i^a + \text{Attn}(\hat{f}_i, f_i^a, f_i^a). 
	\end{equation}
	
	Finally, the output \( f_i^r \) of RCA is then fed into the MHA to further learn long-range dependencies across all positions within the feature space, enhancing the model's representation of point cloud features. MHA can be formalized as follows:
	\begin{equation}
		f_i^g = \text{Attn}(Q_i, K, V) = \text{Softmax}\left(\frac{Q_i K^T}{\sqrt{h}}\right) V,
	\end{equation}
	where \(Q_i = Z \hat{W}^Q\), \(K = Z\hat{W}^K\), and \(V = Z\hat{W}^V\). \(\hat{W}^Q \in \mathbb{R}^{C \times C_h}\), \(\hat{W}^K \in \mathbb{R}^{C \times C_h}\), and \(\hat{W}^V \in \mathbb{R}^{C \times C_h}\) represent the linear mapping matrices. \(Z = \{f_i^r\}_{i=1}^M\) denotes the output matrix of the RCA.

	\subsection{Local Structure Fitting Convolution}
	\label{33}
	
	\subsubsection{Taylor Series}
	
	Inspired by Taylor series, building upon GPM, we adopt a local fitting approach to analyze the local structure and details of the point cloud more finely. The Taylor series is given by:
	
	\begin{equation}
		f(x) = f(a) + \sum_{n=1}^{\infty} \frac{f^{(n)}(a)}{n!}(x - a)^n, \quad |x - a| < \epsilon ,
	\end{equation}
	where \(a\) is a constant, and \(\epsilon\) is an infinitesimal. To simplify the computation, we decompose the Taylor series into a low-frequency component and a high-frequency component, expressed as:
	
	\begin{equation}
		\label{e111}
		f(x) = f(a) + \sum_{n=1}^{\infty} a_n (x - a)^n, \quad |x - a| < \epsilon,
	\end{equation}
	where \(a_n = \frac{f^{(n)}(a)}{n!}\). Based on ~\cref{e111}, it can be understood that, in the representation of local structures within a point cloud, the low-frequency component \(f(a)\) represents the flat parts of the local structure and overall trends of the point cloud, while the high-frequency component \(\sum_{n=1}^{\infty} a_n (x - a)^n\) represents the edges and detailed parts of the local structure.

	\begin{figure}[t]
		\centering
		\includegraphics[width=1.0\linewidth]{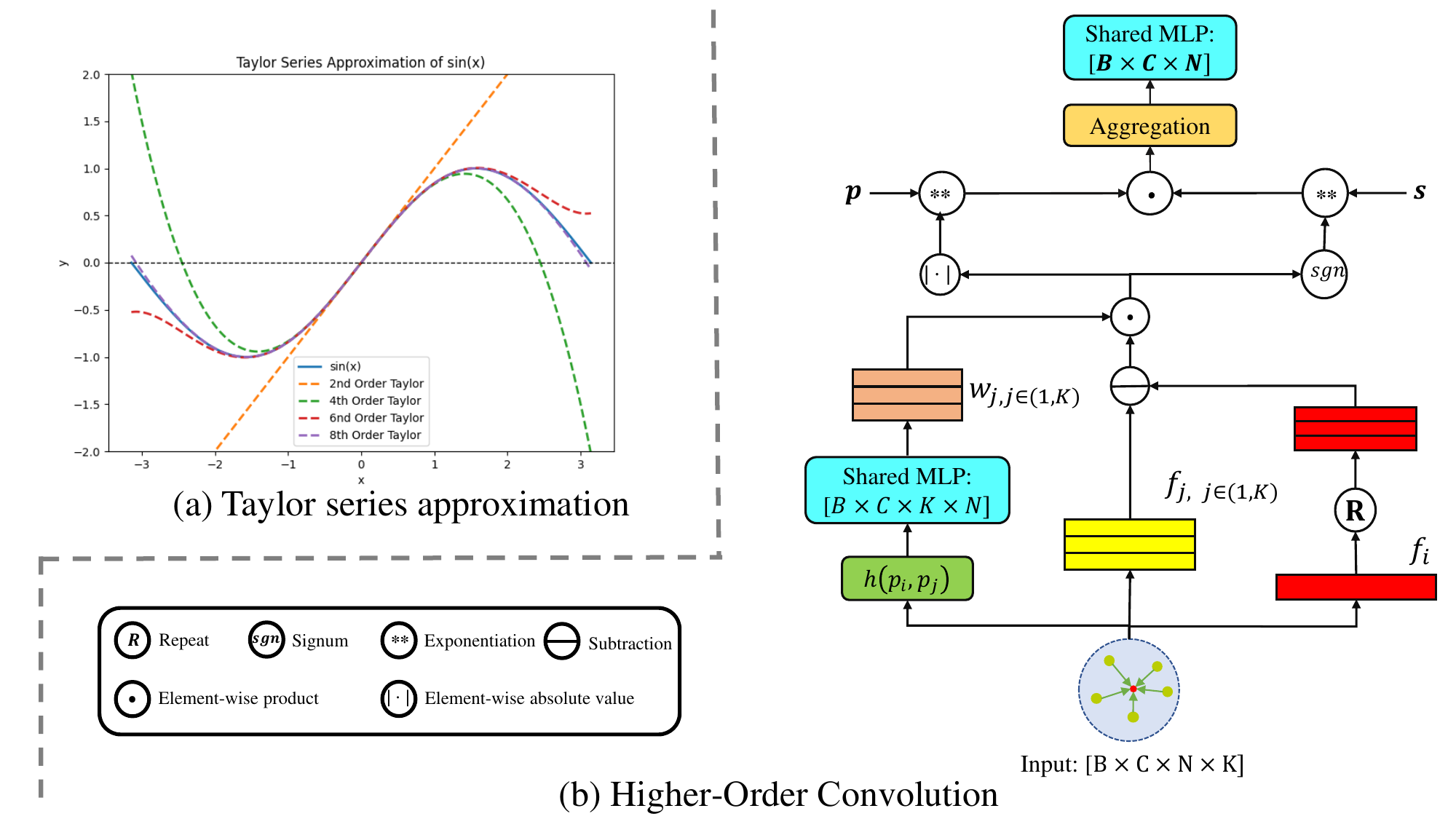}
		\caption{Taylor series and schematic diagram of High-Order Convolution (HOConv).}
		\label{f2}
	\end{figure}
	
	\begin{figure*}[htbp]
		\centering
		\includegraphics[width=1.0\linewidth]{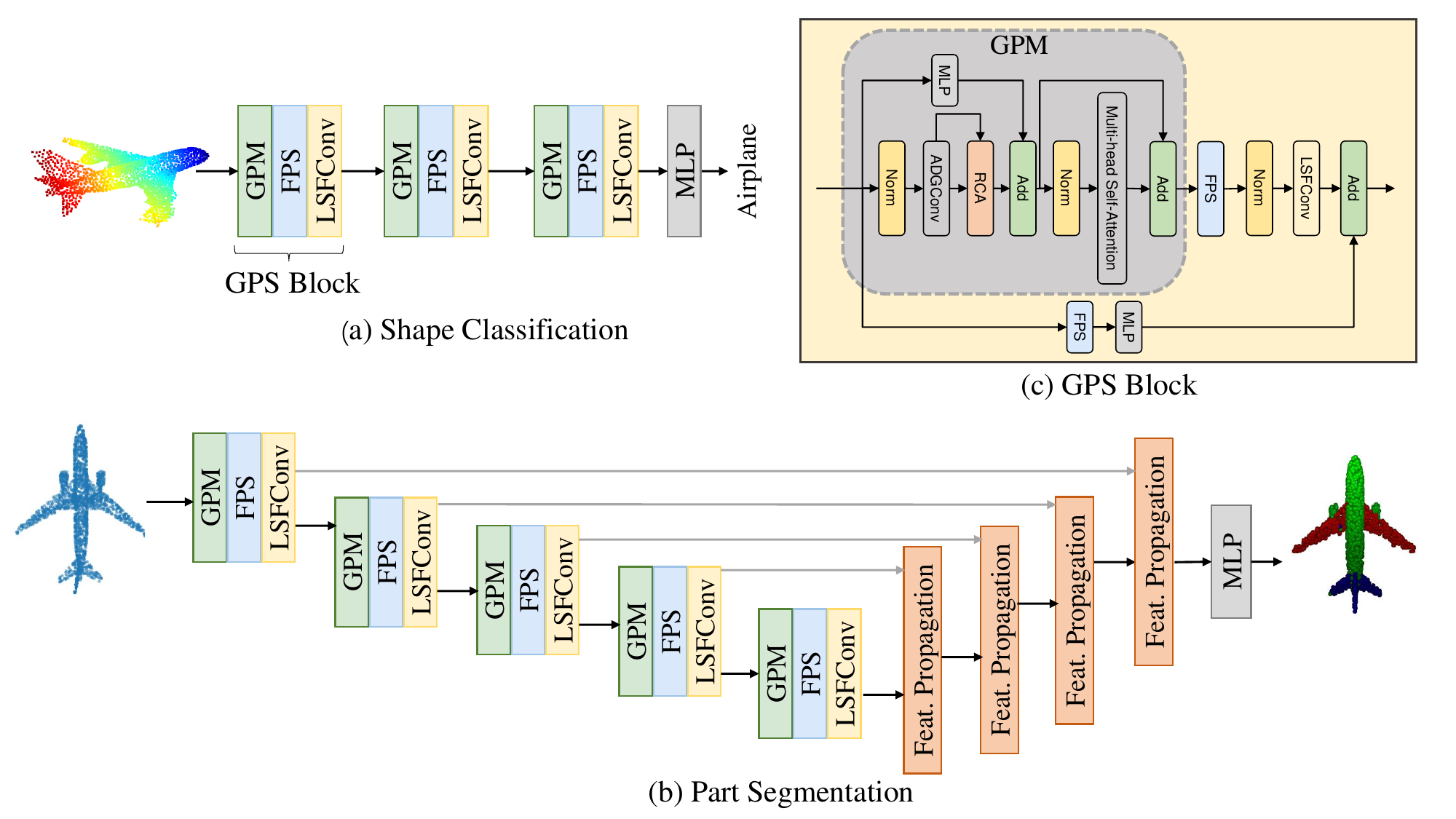}
		
		\caption{Overall Architecture of GPSFormer. For classification (bottom), Three GPSFormer blocks run consecutively, followed by a max-pooling and a multi-layer perceptron. For segmentation (top), a U-net style architecture is adopted with GPSFormer blocks for downsampling and feature propagation for upsampling, followed by a multi-layer perceptron.}
		\label{f3}
	\end{figure*}
	
	\subsubsection{Local Structure Fitting Convolution}
	
	Inspired by Taylor series, we learn both the overall information (low-frequency information) \(f_i^L\) and refined details (high-frequency information) \(f_i^H\) embedded in local structures. Hence, the proposed LSFConv is given by:
	\begin{equation}
		f(\{f_j\}_{j=1}^K) \approx f_i^L + f_i^H = \mathcal{A}(\{\phi(f_j)\}_{j=1}^K) + \mathcal{A}(\{\mathcal{T}(f_i, f_j)\}_{j=1}^K),
	\end{equation}
	\begin{equation}
		\mathcal{T}\left(f_i, f_j\right)=\left(\frac{w_j \cdot\left(f_j-f_i\right)}{\left|w_j \cdot\left(f_j-f_i\right)\right|}\right)^s \cdot\left|w_j \cdot\left(f_j-f_i\right)\right|^p,
	\end{equation}
	where we refer to $f_i^L$ as Low-Order Convolution (LOConv) and $f_i^H$ as High-Order Convolution (HOConv).
	\(\phi\) represents MLP. $\mathcal{T}(f_i, f_j)$ is a novel affine basis function.  Here, \(|\cdot|\) denotes element-wise absolute value, \(s \in \{0,1\}\), and \(p\) is a learnable parameter. When \(s=1, p=1\), \(\mathcal{T}(f_i, f_j)\) degenerates into an Affine Basis Function (ABF~\cite{rosenblatt1958perceptron}) (see \cref{e1}); when \(s=0, p=2\), \(\mathcal{T}(f_i, f_j)\) degenerates into a radial basis function (RBF~\cite{moody1989fast}) (see \cref{e2}). Therefore, the proposed $\mathcal{T}$ exhibits powerful representation capability.
	
	\begin{equation}
		\label{e1}
		\mathcal{T}=\left(\frac{w_j \cdot\left(f_j-0\right)}{\left|w_j \cdot\left(f_j-0\right)\right|}\right)^1 \cdot\left|w_j \cdot\left(f_j-0\right)\right|^1=w_j \cdot f_j,
	\end{equation}
	
	\begin{equation}
		\label{e2}
		\mathcal{T}=\left(\frac{w_j \cdot\left(f_j-f_i\right)}{\left|w_j \cdot\left(f_j-f_i\right)\right|}\right)^0 \cdot\left|w_j \cdot\left(f_j-f_i\right)\right|^2.
	\end{equation}
	
	\subsubsection{Explicit Structure Introduction:} The interaction between sampled points and neighboring points in the point cloud can explicitly reflect the relevance of local point clouds. If we can utilize this prior knowledge to learn weights \(w_j\), it significantly enhances the local structure fitting convolution's ability to perceive the shape of local point clouds. We express the interaction between sampled point \(p_i\) and neighboring point \(p_{ij}\) as:
	
	\begin{equation}
		h\left(p_i, p_{i j}\right)=\left[p_i, p_j, p_j-p_i,\left\|p_i, p_j\right\|\right],
	\end{equation}
	where \(||\cdot||\) denotes the calculation of the Euclidean distance. Therefore, \(w_j\) is defined as:
	\begin{equation}
		w_j = \xi(h(p_i, p_j)),
	\end{equation}
	where \(\xi\) represents MLP.
	
	The explicit introduction of geometric information is beneficial for the local fitting convolution to learn relative spatial layout relationships between points and capture local geometric features and detailed information.
	
	\subsection{GPSFormer}
	\label{34}
	As illustrated in \cref{f3}, building upon the Global Perception Module (GPM) and Local Structure Fitting Convolution (LSFConv), we have designed a model based on the Transformer architecture, termed GPSFormer, for point cloud analysis.
	
	\subsubsection{Point Cloud Classification.} For point cloud classification tasks, we construct a stacked GPSFormer by cascading GPS blocks. In each GPS block, global perception is initially performed utilizing the GPM. Subsequently, representative sampling points are obtained through FPS. Finally, local neighborhoods are constructed around each sampling point, and local shape perception is achieved through the proposed LSFConv. 
	
	We employ three stages of GPS blocks for point cloud classification, with each stage having a feature dimension set of ${64, 128, 256}$. Concurrently, we also evaluate a compact variant, termed GPSFormer-elite, with a feature dimension output set of ${32, 64, 128}$ for each stage. Prediction on the point cloud is carried out through max-pooling and a multi-layer perceptron (MLP). In each stage of LSFConv, a multi-scale strategy is employed for local feature extraction. Spherical queries are utilized to construct local neighborhoods with multi-scale radii ${0.1, 0.2, 0.4}$, corresponding to neighborhood point counts ${8, 16, 32}$. The multi-scale parameters remain consistent across stages, thereby avoiding the hassle of parameter tuning.
	
	\subsubsection{Part Segmentation Task.} For part segmentation, we utilize five stages of GPS blocks in the encoder phase. The method and parameters for constructing local neighborhoods in each stage are identical to those in the classification task. Additionally, a decoder with a reverse interpolation algorithm is employed to restore the point cloud's resolution. Between the encoder and decoder, a skip connection structure similar to U-Net is applied, making full use of contextual information. For more details, please refer to the supplementary material.

	\section{Experiments}
	\label{sec4}
	
	In this section, we demonstrate the effectiveness of the proposed GPSFormer through extensive experiments. First, we conduct experiments on 3D shape classification, part segmentation, and few-shot classfication. Then, we conduct ablation analysis. Finally, we provide visualization to better understand the behavior of GPM and LSFConv.
	
	\subsection{3D Shape Classification}
	\subsubsection{Shape classification on ScanObjectNN.}
	ScanObjectNN~\cite{uy2019revisiting} is a real-world dataset collected for point cloud classification. The objects in this dataset include backgrounds and consider occlusion, making it more realistic and challenging compared to ModelNet40~\cite{wu20153d}. This dataset comprises 2902 point cloud objects, divided into 15 object categories, and we conducted experiments on its most challenging perturbed variant (PB\_T50\_RS).
		\begin{center}
		\begin{minipage}{\textwidth}
			\centering
			\begin{minipage}[t]{0.48\textwidth}
				\makeatletter\def\@captype{table}
				\renewcommand\arraystretch{1.2}
				\caption{Classification results on the ScanObjectNN dataset. “-” denotes unknown. “*” denotes pre-training methods.}
				\label{t1}
				\resizebox{\linewidth}{!}{
					\begin{tabular}{l|c|cc}
						\toprule[2pt]
						Method & Year & mAcc (\%) & OA (\%) \\
						\midrule
						PointNet~\cite{qi2017pointnet} & 2017 & 63.4 & 68.2 \\
						PointNet++~\cite{qi2017pointnet++} & 2018 & 69.8 & 73.7 \\
						PointCNN~\cite{li2018pointcnn} & 2018 & 75.1 & 78.5 \\
						DGCNN~\cite{wang2019dynamic} & 2019 & 73.6 & 78.1 \\
						DRNet~\cite{qiu2021dense} & 2021 & 78.0 & 80.3 \\
						GBNet~\cite{qiu2021geometric} & 2021 & 77.8 & 80.5 \\
						SimpleView~\cite{goyal2021revisiting} & 2021 & - & 80.5 \\
						MVTN~\cite{hamdi2021mvtn} & 2021 & 83.1 & 85.5 \\
						PointMLP~\cite{ma2022rethinking} & 2022 & 84.4 & 85.7 \\
						RepSurf-U~\cite{ran2022surface} & 2022 & 83.1 & 86.0 \\
						PointNeXt~\cite{qian2022pointnext} & 2022 & 85.8 & 87.7 \\
						Point-NN~\cite{zhang2023parameter} & 2023 & - & 64.9 \\
						Point-PN~\cite{zhang2023parameter} & 2023 & - & 87.1 \\
						SPoTr~\cite{park2023self} & 2023 & 86.8 & 88.6 \\
						PointConT~\cite{liu2023point} & 2023 & 88.5 & 90.3 \\
						DeLA~\cite{chen2023decoupled} & 2023 & 89.3 & 90.4 \\
						\midrule
						Point-BERT*~\cite{yu2022point} & 2021 & - & 83.1 \\
						Point-MAE*~\cite{pang2022masked} & 2022 & - & 85.2 \\
						PointFEMAE~\cite{zha2023towards} & 2023 & - & 90.2 \\
						Point-RAE~\cite{liu2023regress} & 2023 & - & 90.3 \\
						ULIP-2 + PointNeXt*~\cite{xue2023ulip} & 2023 & 91.2 & 91.5 \\
						ReCon*~\cite{ren2016recon} & 2023 & - & 91.26 \\
						PointGPT*~\cite{chen2023pointgpt} & 2023 & - & 93.6 \\
						\midrule
						GPSFormer-elite w/o vot. & - & 92.2 & 92.9 \\
						GSPFormer-elite w/vot. & - & 92.5 & 93.3 \\
						GPSFormer w/o vot. & - & 93.5 & 95.0 \\
						GPSFormer w/vot. & - & \textbf{93.8} & \textbf{95.4} \\
						\bottomrule[2pt]
					\end{tabular}
				}
			\end{minipage}
			\hfill
			\begin{minipage}[t]{0.48\textwidth}
				\makeatletter\def\@captype{table}
				\renewcommand
				\arraystretch{1.2}
				\caption{Classification results on ModelNet40 dataset. “-” denotes unknown. “*” denotes pre-training methods.}
				\label{t2}
				\resizebox{\linewidth}{!}{
					\begin{tabular}{l|c|cc}  
						\toprule[2pt]  
						Methods & Year & mAcc (\%) & OA (\%) \\  
						\hline
						
						PointNet~\cite{qi2017pointnet}   &  2016  &   86.0   & 89.2 \\
						PointNet++~\cite{qi2017pointnet++} &  2017  &   -   & 90.7 \\
						PointCNN~\cite{li2018pointcnn}   & 2018   &    -  & 92.2 \\ 
						DGCNN~\cite{wang2019dynamic}      &  2018  &  90.2    & 92.9 \\
						Point Transformer~\cite{zhao2021point} & 2020 & 90.6 & 93.7 \\
						MVTN~\cite{hamdi2021mvtn} &   2020 &   92.2   & 93.8 \\ 
						SimpleView~\cite{goyal2021revisiting} &  2021  &   93.9   & 91.8 \\
						
						PAConv~\cite{xu2021paconv} & 2021 & - & 93.9\\
						
						CurveNet~\cite{muzahid2020curvenet} & 2021 & - & 94.2\\
						
						PointNeXt~\cite{qian2022pointnext} &  2022  &    -  & 94.0 \\ 
						
						PointMLP~\cite{ma2022rethinking} &  2022  &   91.4   & 94.5 \\ 
						RepSurf-U~\cite{ran2022surface} &  2022  &  -    & 94.7 \\ 
						
						Point-NN~\cite{zhang2023parameter} & 2023 &- &81.8  \\
						Point-PN~\cite{zhang2023parameter} & 2023 &- &93.8 \\
						
						PointConT~\cite{liu2023point} & 2023& - & 93.5 \\
						DeLA~\cite{chen2023decoupled} & 2023& 92.2 & 94.0 \\

						\hline
						Point-BERT*~\cite{yu2022point} & 2021 & -& 93.8 \\
						Point-MAE*~\cite{pang2022masked} & 2022 & - & 94.0 \\
						
						PointFEMAE~\cite{zha2023towards}& 2023 & - & 94.5 \\
						Point-RAE~\cite{liu2023regress} & 2023 & - & 94.1 \\

						ULIP + PointMLP*~\cite{xue2023ulip} & 2023 & 92.4 & 94.7 \\
						ReCon*~\cite{ren2016recon} & 2023 & - & 94.7 \\
						PointGPT*~\cite{chen2023pointgpt} &2023 &- &94.9 \\
						
						\hline
						GPSFormer-elite w/o vot.&  -  &   90.9   & 93.4 \\ 
						GPSFormer-elite w/vot. &-   &   91.4   & 93.7 \\ 
						GPSFormer w/o vot. &  -  &   91.8   & 93.8 \\ 
						GPSFormer w/vot.&  -  &  92.2    & 94.2 \\ 
						\bottomrule[2pt]  
					\end{tabular}%
				}
			\end{minipage}
		\end{minipage}
	\end{center}

	We categorize the comparison methods into pure supervised learning and pre-training approaches. \cref{t1} demonstrates that the proposed GPSFormer outperforms all methods, achieving mAcc and OA of 93.8\% and 95.4\% respectively. This result is 1.8\% higher than the pre-training method PointGPT~\cite{chen2023pointgpt}'s OA, and 4.5\% and 5.0\% higher than the pure supervised method DeLA~\cite{chen2023decoupled}'s mAcc and OA, respectively. Even the compact GSPFormer-elite achieved OA of 93.3\% with only 0.68M parameters. This outcome showcases GPSFormer's robust ability to capture long-range dependencies and local structural representations.

	\subsubsection{Shape classification on ModelNet40.}

	The ModelNet40~\cite{wu20153d} dataset, widely regarded as the benchmark for point cloud analysis, consists of point clouds 
	\begin{wraptable}{r}{0.5\textwidth}
		\renewcommand
		\arraystretch{1.2}
		\centering  
		\caption{Part segmentation results (\%) on the ShapeNetPart. Mean IoU of all part categories (class mIoU) and mean IoU of all instances (instance mIoU) are reported.}    
		
		\resizebox{63mm}{!}{    
			\begin{tabular}{l|c|cc}    
				\toprule[2pt]    
				Methods & Year & class mIoU & instance mIoU \\    
				\hline    
				
				PointNet~\cite{qi2017pointnet} & 2017 & 80.4 & 83.7 \\  
				PointNet++~\cite{qi2017pointnet++} & 2017 & 81.9 & 85.1 \\  
				PointCNN~\cite{li2018pointcnn}  & 2018 & 84.6 & 86.1 \\  
				DGCNN~\cite{wang2019dynamic} & 2019 & 82.3 & 85.1 \\  
				RS-CNN~\cite{liu2019relation}  & 2019 & 84.0 & 86.2 \\  
				KPConv~\cite{thomas2019kpconv} & 2019 & 85.1 & 86.4 \\  
				PointConv~\cite{wu2019pointconv} & 2019 & 82.8 & 85.7 \\  
				Point Transformer~\cite{zhao2021point} & 2020 & - & 85.9\\  
				PointASNL~\cite{yan2020pointasnl} & 2020 & - & 86.1 \\  
				PCT~\cite{guo2021pct} & 2021 & - & 86.4 \\  
				PAConv~\cite{xu2021paconv}  & 2021 & 84.6 & 86.1 \\  
				AdaptConv~\cite{zhou2021adaptive} & 2021 & 83.4 & 86.4 \\  
				Point Transformer~\cite{zhao2021point} & 2021 & 83.7 & 86.6 \\  
				CurveNet~\cite{muzahid2020curvenet}  & 2021 & - & 86.8 \\  
				PointMLP~\cite{ma2022rethinking}  & 2022 & 84.6 & 86.1 \\  
				PointNeXt~\cite{qian2022pointnext} & 2022 & 85.2  & 87.1  \\  
				SPoTr~\cite{park2023self} & 2023 & 85.4 & 87.2 \\  
				\hline    
				GPSFormer &  -  & 85.4   & 86.8\\   
				\bottomrule[2pt]    
			\end{tabular}%
			\label{t3}     
		}    
	\end{wraptable}
	representing composite objects. It encompasses 40 categories (such as aircraft, cars, plants, and lights), with 9843 samples used for training and the remaining 2468 for testing.

	As shown in \cref{t2}, the proposed GPSFormer achieves an impressive accuracy of 94.2\% on the synthetic ModelNet40 dataset, outperforming most supervised and pre-training methods. This outcome underscores the effectiveness and generalizability of GPSFormer. Currently, existing methods have essentially reached saturation on the less challenging synthetic ModelNet40 dataset, with a performance gap of less than 0.8\% among these advanced approaches. However, this marginal gap belies their relatively poor performance on real-world datasets like ScanObjectNN. Consequently, the ModelNet40 dataset alone cannot serve as an accurate evaluation benchmark for model performance.

	\subsection{Part Segmentation}

	ShapeNetPart~\cite{yi2016scalable} is a subset of the large-scale 3D CAD template library ShapeNet, which contains 16,881 shapes of 16 common object classes (i.e., table, chair, plane, etc.). Each shape is annotated with 2-5 parts, resulting in a total of 50 part categories in the dataset. In this experiment, we used 13,807 models for training and 2,874 models for testing.

	As shown in \cref{t3}, we evaluated the performance of GPSFormer for part segmentation using the mean IoU and class IoU of each instance. We randomly selected 2048 points as input and reported the results after voting for 10 times. Obviously, compared with existing methods, GPSFomer achieved competitive results, especially achieving the best performance of 85.4\% in terms of class IoU. \cref{f4} shows the results of some part segmentation, indicating that GPSFormer can effectively recognize the shape information of objects and accurately partition similar components.

	\subsection{Few-Shot Classification}
	
	The existing methods perform few-shot classification on ModelNet40~\cite{wu20153d}. To better reflect the model's ability in complex environments, we provide a few-shot dataset for ScanObjectNN~\cite{uy2019revisiting} following the division method of ModelNet40. According to the setting of previous works, we sampled the "$n$-way $m$-shot" setting, which randomly selects $n$ classes from the dataset and randomly selects $m$ samples from each class for training. During testing, 20 samples are randomly selected from the remaining samples of n classes. Therefore, we evaluated in four settings (5-way 10-shot, 5-way 20-shot, 10-way 10-shot, 10-way 20-shot), each setting conducted 10 independent experiments, and the average value of the experimental results was used as the performance indicator of the model. \cref{t4} and \cref{t5} provide the few-shot classification performance of GPSFormer on ModelNet40 and ScanObjectNN. It can be seen that the proposed GPSFormer can learn robust shape information in limited samples.

\begin{center}
	\begin{minipage}{\textwidth}
		\centering
		\begin{minipage}[t]{0.48\textwidth}
			\makeatletter\def\@captype{table}
			\renewcommand\arraystretch{1.2}
			\caption{Few-shot classification results on the ModelNet40 dataset. Mean accuracy (\%) and standard deviation are reported across 10 independent trials for each scenario.}
			\label{t4}
			\resizebox{\linewidth}{!}{
				\begin{tabular}{lccccc}  
					\hline  
					& \multicolumn{2}{c}{5-way} & & \multicolumn{2}{c}{10-way} \\  
					\cline{2-3} \cline{5-6}   
					& 10-shot & 20-shot & & 10-shot & 20-shot \\  
					\hline   
					DGCNN~\cite{wang2019dynamic} & 31.6 & 40.8 & & 19.9 & 16.9 \\
					FoldingNet~\cite{yang2018foldingnet} & 33.4 & 35.8 & & 18.6 & 15.4 \\
					PointNet++~\cite{qi2017pointnet++} & 38.5 & 42.4 & & 23.0 & 18.8 \\
					PointNet~\cite{qi2017pointnet} & 52.0 & 57.8 & & 46.6 & 35.2 \\
					3D-GAN~\cite{wu2016learning} & 55.8 & 65.8 & & 40.3 & 48.4 \\
					PointCNN~\cite{li2018pointcnn} & 65.4 & 68.6 & & 46.6 & 50.0 \\
					Point-NN~\cite{zhang2023parameter} & 88.8 & 90.9 & & 79.9 & 84.9 \\
					\hline 
					GPSFormer & 90.1 & 91.5 & & 82.3 & 86.2 \\
					\hline  
				\end{tabular}%
			}
		\end{minipage}
		\hfill
		\begin{minipage}[t]{0.48\textwidth}
			\makeatletter\def\@captype{table}
			\renewcommand\arraystretch{1.27}
			\caption{Impact of each component of GPM on the ScanObjectNN.}  
			\label{t6} 
			\resizebox{\linewidth}{!}{
				\begin{tabular}{ccc|c}  
					\hline  
					\multicolumn{3}{c|}{ Settings } & \multirow{2}{*}{ OA(\%) } \\
					\cline { 1 - 3 } ADGConv & RCA & MHA & \\
					\hline$\checkmark$ & & & 93.2 \\
					& $\checkmark$ & & 88.7 \\
					& & $\checkmark$ & 89.6 \\
					$\checkmark$ & $\checkmark$ & & 94.4 \\
					$\checkmark$ & & $\checkmark$ & 95.0 \\
					$\checkmark$ & $\checkmark$ & $\checkmark$ & $\mathbf{95.4}$ \\
					\hline  
				\end{tabular}%
			}
		\end{minipage}

	\end{minipage}
\end{center}

\begin{center}
	\begin{minipage}{\textwidth}
		\centering
		\begin{minipage}[t]{0.48\textwidth}
			\makeatletter\def\@captype{table}
			\renewcommand\arraystretch{1.32}
			\caption{Few-shot classification on ScanObjectNN. $^\dagger$ denotes a model trained from scratch.}
			\label{t5}
			\resizebox{\linewidth}{!}{
				\begin{tabular}{lccccc}  
					\hline  
					& \multicolumn{2}{c}{5-way} & & \multicolumn{2}{c}{10-way} \\  
					\cline {2-3} \cline {5-6}   
					& 10-shot & 20-shot & & 10-shot & 20-shot \\  
					\hline   
					PointNeXt$^\dagger$\cite{qian2022pointnext} & 55.7 & 53.4 & & 39.6 & 42.8 \\
					PointNeXt~\cite{qian2022pointnext} & 72.4 & 72.2 & & 68.9 & 69.5 \\
					\hline 
					GPSFormer$^\dagger$ & 71.7 & 73.6 & & 54.3 & 62.1 \\
					GPSFormer & 89.3 & 87.0 & & 86.6 & 87.0 \\
					\hline  
				\end{tabular}%
			}
		\end{minipage}
		\hfill
		\begin{minipage}[t]{0.48\textwidth}
			\makeatletter\def\@captype{table}
			\renewcommand\arraystretch{1.2}
			\caption{The Influence of HOConv's Parameters on GPSFormer.}
			\label{t8}
			\resizebox{\linewidth}{!}{
				\begin{tabular}{c|c}
				\toprule[1pt]
				parameter settings & Accuracy (\%)  \\ \hline
				ABF~\cite{rosenblatt1958perceptron} & 92.8  \\ 
				RBF~\cite{moody1989fast} & 93.2  \\ 
				s=0,$\mathrm{p}$ learnable  & 94.6  \\ 
				s=1,$\mathrm{p}$ learnable  & $\mathbf{95.4}$   \\
				\bottomrule[1pt]
			\end{tabular}		
			}
		\end{minipage}
		
	\end{minipage}
\end{center}

		\subsection{Ablation Study}
		
		\subsubsection{The Effectiveness of the Global Perception Module.} As shown in \cref{t6}, it is evident that ADGConv, RCA, and MHA modules play crucial roles in point cloud context modeling. Individually, ADGConv effectively extracts local features through dynamic graph convolution, achieving 93.2\% accuracy. RCA contributes to global relationships with 88.7\% accuracy, while MHA captures dependencies, though less effectively than ADGConv and RCA. Combining ADGConv and RCA improves accuracy to 94.4\%, highlighting their complementarity. The combination of ADGConv and MHA achieves 95.0\%, validating their effectiveness. Finally, utilizing all three modules together attains the highest accuracy of 95.4\%, emphasizing their synergistic roles in enhancing model performance for point cloud analysis.
		
		\begin{center}
	\begin{minipage}{\textwidth}
		\centering
		\begin{minipage}[t]{0.48\textwidth}
			\makeatletter\def\@captype{table}
			\renewcommand\arraystretch{1.2}
			\caption{The Influence of the neighborhood radius of ADGConv.} 
			\label{t7} 
			\resizebox{60mm}{!}{  
				\begin{tabular}{c|c}  
					\hline  
					\text{The number of neighbor points}  & OA(\%)\\
					\hline  
					5 & 88.6\\
					10 & 92.4\\
					15 & 94.3 \\
					20 & $\mathbf{95.4}$ \\
					25 & 94.8 \\
					30 & 93.2\\
					\hline  
				\end{tabular}%
			}
		\end{minipage}
		\hfill
		\begin{minipage}[t]{0.48\textwidth}
			\makeatletter\def\@captype{table}
			\renewcommand\arraystretch{1.18}
			\caption{Model complexity comparison on ScanObjectNN.}    
			\resizebox{60mm}{!}{    
				\begin{tabular}{l|c cc}    
					\toprule[2pt]    
					Methods & OA(\%) & parameter(M) &  FLOPS(G) \\    
					\hline    
					PointNet~\cite{qi2017pointnet} & 68.0 & 3.5  & 0.5 \\  
					PointNet++~\cite{qi2017pointnet++} & 77.9 & 1.5  & 1.7 \\  
					DGCNN~\cite{wang2019dynamic} & 78.1 & 1.8  & 2.4 \\  
					PointCNN~\cite{li2018pointcnn}  & 78.5 & 0.6 & - \\  
					MVTN~\cite{hamdi2021mvtn}  & 82.8 & 11.2 & 43.7 \\  
					PointMLP~\cite{ma2022rethinking}  & 85.4 & 12.6 & 31.4 \\  
					PointNeXt~\cite{qian2022pointnext} & 87.4 & 1.4  & 3.6  \\  
					\hline    
					GPSFormer &  95.4  & 2.36   & 0.7\\   
					\bottomrule[2pt]    
				\end{tabular}%
				\label{t9}  		
			}
		\end{minipage}
		
	\end{minipage}
\end{center}
		
		\subsubsection{Neighborhood size of ADGConv.} Since ADGConv aggregates features within feature neighborhoods, neighboring points have strong semantic relationships in spatial positions. If the neighborhood is too small, it will cause ADGConv to become local for feature aggregation within the neighborhood; if the neighborhood is too large, it not only increases the search time of the model but also introduces some irrelevant features of points. As shown in Table \ref{t7}, when the size of the neighborhood is 20, the model can learn good contextual information. However, when the neighborhood is too large or too small, valuable semantic information will not be learned.

		\subsubsection{The Parameter Influence of High Order Convolution.} The High-Order Convolution (HOConv) plays a critical role in shaping the network's performance, as demonstrated by the results in \cref{t8}. The configuration with a learnable parameter $p$ and setting $s=1$ emerged as the most efficacious, achieving a remarkable accuracy of 95.4\%. This highlights the significance of adaptive parameter tuning in enhancing network outcomes. Although the baseline methodologies ABF~\cite{rosenblatt1958perceptron} and RBF~\cite{moody1989fast} recorded commendable accuracies of 92.8\% and 93.2\% respectively, the adaptive parameter settings clearly surpassed them.
		
	\begin{figure}[tb]
	\centering
	\begin{subfigure}{0.48\linewidth}
		\centering
		\includegraphics[width=1.0\linewidth]{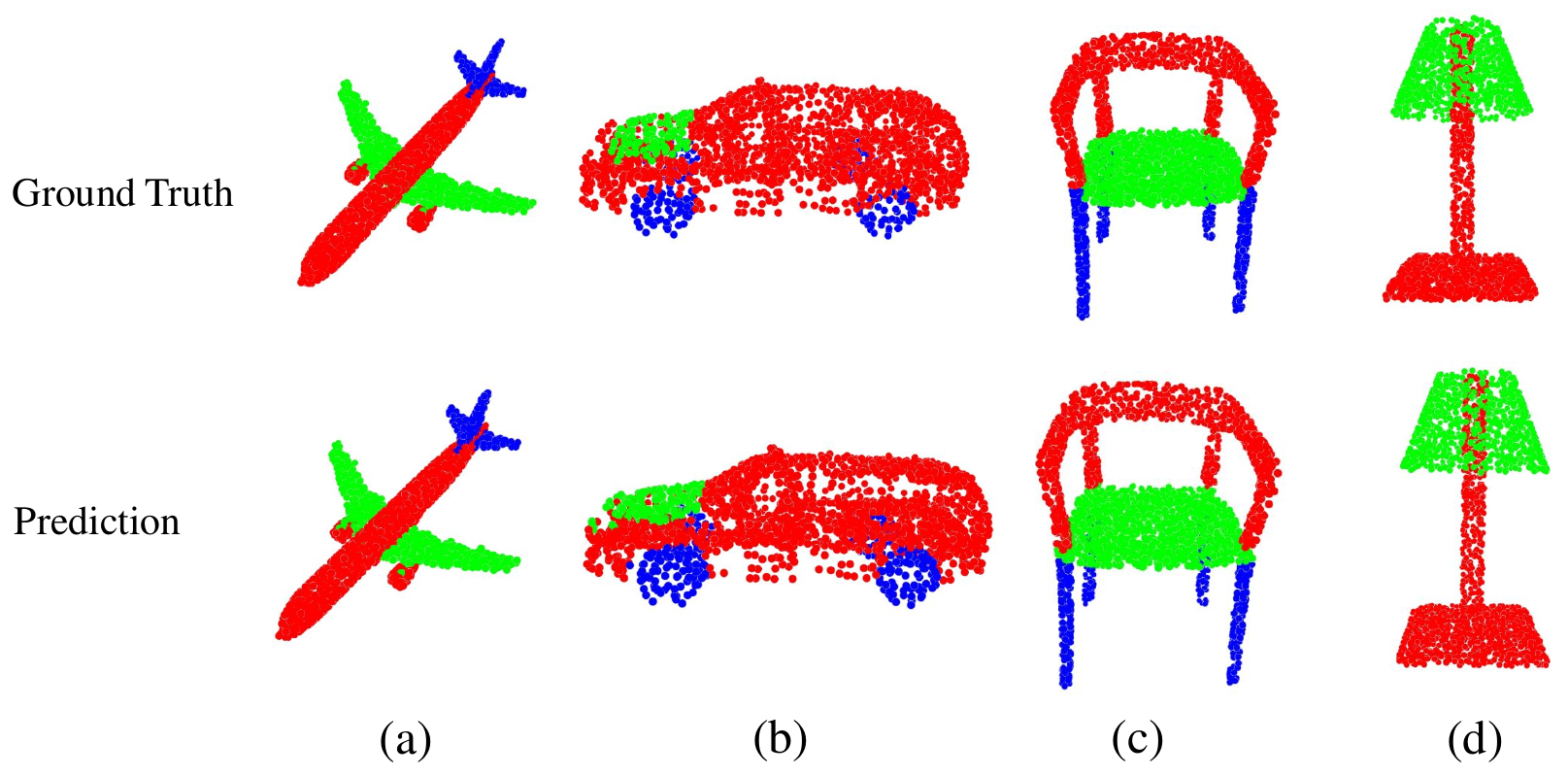}
		\caption{}
		\label{f4}
	\end{subfigure}
	\hfill
	\begin{subfigure}{0.5\linewidth}
		\centering
		\includegraphics[width=1.0\linewidth]{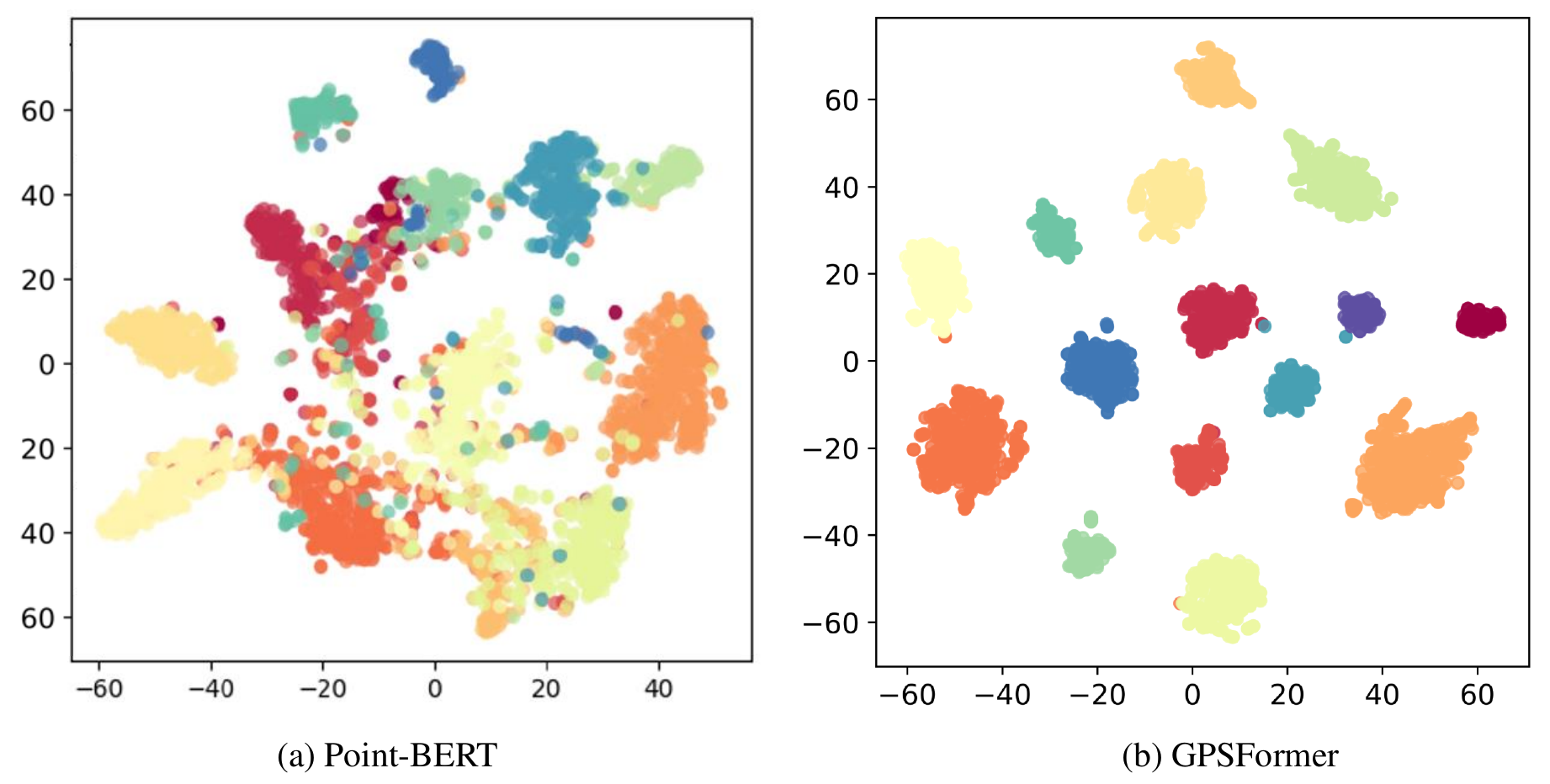}
		\caption{}
		\label{f5}
	\end{subfigure}
	\caption{(a) Visualization of object part segmentation results on the ShapeNetPart dataset. The ground truth is shown in the top row, with predictions below. (b) t-SNE visualization comparing the feature spaces of Point-BERT~\cite{yu2022point} and GPSFormer on the ScanObjectNN dataset.}
	\label{fig:56}
\end{figure}

		\subsection{Visualization}
		\cref{f5} visualizes the spatial distribution of sample features on the ScanObjectNN dataset using Point-Bert~\cite{yu2022point} and GPSFormer. It can be seen that the proposed GPSFormer can better reduce the within-class distance of objects and make the sample distribution more compact, effectively recognizing the shape information of objects.

		\subsection{Model Complexity}
		\cref{t9} provides an evaluation of the model complexity of GPSFormer on the ScanObjectNN dataset. GPSFormer achieves the best results in terms of speed and accuracy with only a slight increase in parameter count, featuring just 2.36M parameters and 0.7G FLOPS. This highlights its effectiveness and efficiency in point cloud understanding tasks.

		\section{Conclusion}
		
		In this paper, we propose a novel \textbf{G}lobal \textbf{P}erception and Local \textbf{S}tructure \textbf{F}itting-based Transf\textbf{ormer} (\textbf{GPSFormer}) to address the challenge of effectively capturing shape information from irregular point clouds. The key contributions of GPSFormer include introducing a Global Perception Module (GPM) and a Local Structure Fitting Convolution (LSFConv). The GPM enhances the model's ability to capture global contextual information in point clouds by introducing the Adaptive Deformable Graph Convolution (ADGConv). Meanwhile, the LSFConv finely learns the local geometric structures of point clouds, acquiring both low-order fundamental information and high-order refinement details. Through extensive experiments, the proposed GPSFormer demonstrates efficient processing and analysis of point clouds without relying on external data. In the future, we plan to further explore the potential of GPSFormer in pre-training, lightweight approaches, and few-shot learning settings.

		\section*{Acknowledgment}
		
		This work was supported in part by NTU-DESAY SV Research Program under Grant 2018-0980; and in part by the Ministry of Education, Singapore, under its Academic Research Fund Tier 1, under Grant RG78/21. The computational work for this article was partially performed on resources of the National Supercomputing Centre, Singapore (\url{https://www.nscc.sg}).
		
		\bibliographystyle{splncs04}
		\bibliography{reference}

\begin{thebibliography}{10}
\providecommand{\url}[1]{\texttt{#1}}
\providecommand{\urlprefix}{URL }
\providecommand{\doi}[1]{https://doi.org/#1}

\bibitem{chen2023decoupled}
Chen, B., Xia, Y., Zang, Y., Wang, C., Li, J.: Decoupled local aggregation for
  point cloud learning. arXiv preprint arXiv:2308.16532  (2023)

\bibitem{chen2023pointgpt}
Chen, G., Wang, M., Yang, Y., Yu, K., Yuan, L., Yue, Y.: Pointgpt:
  Auto-regressively generative pre-training from point clouds. arXiv preprint
  arXiv:2305.11487  (2023)

\bibitem{engel2021point}
Engel, N., Belagiannis, V., Dietmayer, K.: Point transformer. IEEE Access
  \textbf{9},  134826--134840 (2021)

\bibitem{fang2021animc}
Fang, X., Hu, Y., Zhou, P., Wu, D.: Animc: A soft approach for autoweighted
  noisy and incomplete multiview clustering. IEEE Transactions on Artificial
  Intelligence  \textbf{3}(2),  192--206 (2021)

\bibitem{fang2020v}
Fang, X., Hu, Y., Zhou, P., Wu, D.O.: V3h: View variation and view heredity for
  incomplete multiview clustering. IEEE Transactions on Artificial Intelligence
   \textbf{1}(3),  233--247 (2020)

\bibitem{fang2021unbalanced}
Fang, X., Hu, Y., Zhou, P., Wu, D.O.: Unbalanced incomplete multi-view
  clustering via the scheme of view evolution: Weak views are meat; strong
  views do eat. IEEE Transactions on Emerging Topics in Computational
  Intelligence  \textbf{6}(4),  913--927 (2021)

\bibitem{fang2023annotations}
Fang, X., Liu, D., Fang, W., Zhou, P., Cheng, Y., Tang, K., Zou, K.:
  Annotations are not all you need: A cross-modal knowledge transfer network
  for unsupervised temporal sentence grounding. In: Findings of the Association
  for Computational Linguistics: EMNLP 2023. pp. 8721--8733 (2023)

\bibitem{fang2024fewer}
Fang, X., Liu, D., Fang, W., Zhou, P., Xu, Z., Xu, W., Chen, J., Li, R.: Fewer
  steps, better performance: Efficient cross-modal clip trimming for video
  moment retrieval using language. In: Proceedings of the AAAI Conference on
  Artificial Intelligence. pp. 1735--1743 (2024)

\bibitem{fang2022multi}
Fang, X., Liu, D., Zhou, P., Hu, Y.: Multi-modal cross-domain alignment network
  for video moment retrieval. IEEE Transactions on Multimedia  \textbf{25},
  7517--7532 (2022)

\bibitem{fang2023you}
Fang, X., Liu, D., Zhou, P., Nan, G.: You can ground earlier than see: An
  effective and efficient pipeline for temporal sentence grounding in
  compressed videos. In: Proceedings of the IEEE/CVF Conference on Computer
  Vision and Pattern Recognition. pp. 2448--2460 (2023)

\bibitem{fang2023hierarchical}
Fang, X., Liu, D., Zhou, P., Xu, Z., Li, R.: Hierarchical local-global
  transformer for temporal sentence grounding. IEEE Transactions on Multimedia
  (2023)

\bibitem{goyal2021revisiting}
Goyal, A., Law, H., Liu, B., Newell, A., Deng, J.: Revisiting point cloud shape
  classification with a simple and effective baseline. In: International
  Conference on Machine Learning. pp. 3809--3820. PMLR (2021)

\bibitem{guo2021pct}
Guo, M.H., Cai, J.X., Liu, Z.N., Mu, T.J., Martin, R.R., Hu, S.M.: Pct: Point
  cloud transformer. Computational Visual Media  \textbf{7},  187--199 (2021)

\bibitem{hamdi2021mvtn}
Hamdi, A., Giancola, S., Ghanem, B.: Mvtn: Multi-view transformation network
  for 3d shape recognition. In: ICCV. pp. 1--11 (2021)

\bibitem{jiang2023lttpoint}
Jiang, L., Wang, C., Ning, X., Yu, Z.: Lttpoint: A mlp-based point cloud
  classification method with local topology transformation module. In: 2023 7th
  Asian Conference on Artificial Intelligence Technology (ACAIT). pp. 783--789.
  IEEE (2023)

\bibitem{kasneci2023chatgpt}
Kasneci, E., Se{\ss}ler, K., K{\"u}chemann, S., Bannert, M., Dementieva, D.,
  Fischer, F., Gasser, U., Groh, G., G{\"u}nnemann, S., H{\"u}llermeier, E.,
  et~al.: Chatgpt for good? on opportunities and challenges of large language
  models for education. Learning and individual differences  \textbf{103},
  102274 (2023)

\bibitem{komarichev2019cnn}
Komarichev, A., Zhong, Z., Hua, J.: A-cnn: Annularly convolutional neural
  networks on point clouds. In: CVPR. pp. 7421--7430 (2019)

\bibitem{li2020end}
Li, L., Zhu, S., Fu, H., Tan, P., Tai, C.L.: End-to-end learning local
  multi-view descriptors for 3d point clouds. In: CVPR. pp. 1919--1928 (2020)

\bibitem{li2018pointcnn}
Li, Y., Bu, R., Sun, M., Wu, W., Di, X., Chen, B.: Pointcnn: Convolution on
  x-transformed points. NeurIPS  \textbf{31} (2018)

\bibitem{liu2023point}
Liu, Y., Tian, B., Lv, Y., Li, L., Wang, F.Y.: Point cloud classification using
  content-based transformer via clustering in feature space. IEEE/CAA Journal
  of Automatica Sinica  (2023)

\bibitem{liu2023regress}
Liu, Y., Chen, C., Wang, C., King, X., Liu, M.: Regress before construct:
  Regress autoencoder for point cloud self-supervised learning. In: ACMMM. pp.
  1738--1749 (2023)

\bibitem{liu2019relation}
Liu, Y., Fan, B., Xiang, S., Pan, C.: Relation-shape convolutional neural
  network for point cloud analysis. In: CVPR. pp. 8895--8904 (2019)

\bibitem{ma2022rethinking}
Ma, X., Qin, C., You, H., Ran, H., Fu, Y.: Rethinking network design and local
  geometry in point cloud: A simple residual mlp framework. arXiv preprint
  arXiv:2202.07123  (2022)

\bibitem{maturana2015voxnet}
Maturana, D., Scherer, S.: Voxnet: A 3d convolutional neural network for
  real-time object recognition. In: 2015 IEEE/RSJ International Conference on
  Intelligent Robots and Systems (IROS). pp. 922--928. IEEE (2015)

\bibitem{moody1989fast}
Moody, J., Darken, C.J.: Fast learning in networks of locally-tuned processing
  units. Neural computation  \textbf{1}(2),  281--294 (1989)

\bibitem{muzahid2020curvenet}
Muzahid, A., Wan, W., Sohel, F., Wu, L., Hou, L.: Curvenet: Curvature-based
  multitask learning deep networks for 3d object recognition. IEEE/CAA Journal
  of Automatica Sinica  \textbf{8}(6),  1177--1187 (2020)

\bibitem{pang2022masked}
Pang, Y., Wang, W., Tay, F.E., Liu, W., Tian, Y., Yuan, L.: Masked autoencoders
  for point cloud self-supervised learning. In: ECCV. pp. 604--621. Springer
  (2022)

\bibitem{park2023self}
Park, J., Lee, S., Kim, S., Xiong, Y., Kim, H.J.: Self-positioning point-based
  transformer for point cloud understanding. In: CVPR. pp. 21814--21823 (2023)

\bibitem{qi2017pointnet}
Qi, C.R., Su, H., Mo, K., Guibas, L.J.: Pointnet: Deep learning on point sets
  for 3d classification and segmentation. In: CVPR. pp. 652--660 (2017)

\bibitem{qi2017pointnet++}
Qi, C.R., Yi, L., Su, H., Guibas, L.J.: Pointnet++: Deep hierarchical feature
  learning on point sets in a metric space. arXiv preprint arXiv:1706.02413
  (2017)

\bibitem{qian2022pointnext}
Qian, G., Li, Y., Peng, H., Mai, J., Hammoud, H., Elhoseiny, M., Ghanem, B.:
  Pointnext: Revisiting pointnet++ with improved training and scaling
  strategies. NeurIPS  \textbf{35},  23192--23204 (2022)

\bibitem{qiu2021dense}
Qiu, S., Anwar, S., Barnes, N.: Dense-resolution network for point cloud
  classification and segmentation. In: Proceedings of the IEEE/CVF Winter
  Conference on Applications of Computer Vision. pp. 3813--3822 (2021)

\bibitem{qiu2021geometric}
Qiu, S., Anwar, S., Barnes, N.: Geometric back-projection network for point
  cloud classification. TMM  \textbf{24},  1943--1955 (2021)

\bibitem{ran2022surface}
Ran, H., Liu, J., Wang, C.: Surface representation for point clouds. In: CVPR.
  pp. 18942--18952 (2022)

\bibitem{rao2020global}
Rao, Y., Lu, J., Zhou, J.: Global-local bidirectional reasoning for
  unsupervised representation learning of 3d point clouds. In: CVPR. pp.
  5376--5385 (2020)

\bibitem{ren2016recon}
Ren, J., Rao, A., Lindorfer, M., Legout, A., Choffnes, D.: Recon: Revealing and
  controlling pii leaks in mobile network traffic. In: Proceedings of the 14th
  Annual International Conference on Mobile Systems, Applications, and
  Services. pp. 361--374 (2016)

\bibitem{riegler2017octnet}
Riegler, G., Osman~Ulusoy, A., Geiger, A.: Octnet: Learning deep 3d
  representations at high resolutions. In: CVPR. pp. 3577--3586 (2017)

\bibitem{rosenblatt1958perceptron}
Rosenblatt, F.: The perceptron: a probabilistic model for information storage
  and organization in the brain. Psychological review  \textbf{65}(6), ~386
  (1958)

\bibitem{su2015multi}
Su, H., Maji, S., Kalogerakis, E., Learned-Miller, E.: Multi-view convolutional
  neural networks for 3d shape recognition. In: ICCV. pp. 945--953 (2015)

\bibitem{sun2020scalability}
Sun, P., Kretzschmar, H., Dotiwalla, X., Chouard, A., Patnaik, V., Tsui, P.,
  Guo, J., Zhou, Y., Chai, Y., Caine, B., et~al.: Scalability in perception for
  autonomous driving: Waymo open dataset. In: CVPR. pp. 2446--2454 (2020)

\bibitem{thomas2019kpconv}
Thomas, H., Qi, C.R., Deschaud, J.E., Marcotegui, B., Goulette, F., Guibas,
  L.J.: Kpconv: Flexible and deformable convolution for point clouds. In: ICCV.
  pp. 6411--6420 (2019)

\bibitem{uy2019revisiting}
Uy, M.A., Pham, Q.H., Hua, B.S., Nguyen, T., Yeung, S.K.: Revisiting point
  cloud classification: A new benchmark dataset and classification model on
  real-world data. In: ICCV. pp. 1588--1597 (2019)

\bibitem{wang20233d}
Wang, C., Ning, X., Li, W., Bai, X., Gao, X.: 3d person re-identification based
  on global semantic guidance and local feature aggregation. IEEE Transactions
  on Circuits and Systems for Video Technology  (2023)

\bibitem{wang2022learning}
Wang, C., Ning, X., Sun, L., Zhang, L., Li, W., Bai, X.: Learning
  discriminative features by covering local geometric space for point cloud
  analysis. IEEE Transactions on Geoscience and Remote Sensing  \textbf{60},
  1--15 (2022)

\bibitem{wangchangshuo20223d}
Wang, C., Wang, H., Ning, X., Shengwei, T., Li, W.: 3d point cloud
  classification method based on dynamic coverage of local area. Journal of
  Software  \textbf{34}(4),  1962--1976 (2022)

\bibitem{wang2018local}
Wang, C., Samari, B., Siddiqi, K.: Local spectral graph convolution for point
  set feature learning. In: ECCV. pp. 52--66 (2018)

\bibitem{wang2019dynamic}
Wang, Y., Sun, Y., Liu, Z., Sarma, S.E., Bronstein, M.M., Solomon, J.M.:
  Dynamic graph cnn for learning on point clouds. Acm Transactions On Graphics
  (tog)  \textbf{38}(5),  1--12 (2019)

\bibitem{wei2020view}
Wei, X., Yu, R., Sun, J.: View-gcn: View-based graph convolutional network for
  3d shape analysis. In: CVPR. pp. 1850--1859 (2020)

\bibitem{wu2016learning}
Wu, J., Zhang, C., Xue, T., Freeman, B., Tenenbaum, J.: Learning a
  probabilistic latent space of object shapes via 3d generative-adversarial
  modeling. NeurIPS  \textbf{29} (2016)

\bibitem{wu2019pointconv}
Wu, W., Qi, Z., Fuxin, L.: Pointconv: Deep convolutional networks on 3d point
  clouds. In: CVPR. pp. 9621--9630 (2019)

\bibitem{wu20153d}
Wu, Z., Song, S., Khosla, A., Yu, F., Zhang, L., Tang, X., Xiao, J.: 3d
  shapenets: A deep representation for volumetric shapes. In: CVPR. pp.
  1912--1920 (2015)

\bibitem{xu2021paconv}
Xu, M., Ding, R., Zhao, H., Qi, X.: Paconv: Position adaptive convolution with
  dynamic kernel assembling on point clouds. In: CVPR. pp. 3173--3182 (2021)

\bibitem{xue2023ulip}
Xue, L., Gao, M., Xing, C., Mart{\'\i}n-Mart{\'\i}n, R., Wu, J., Xiong, C., Xu,
  R., Niebles, J.C., Savarese, S.: Ulip: Learning a unified representation of
  language, images, and point clouds for 3d understanding. In: CVPR. pp.
  1179--1189 (2023)

\bibitem{yan2020pointasnl}
Yan, X., Zheng, C., Li, Z., Wang, S., Cui, S.: Pointasnl: Robust point clouds
  processing using nonlocal neural networks with adaptive sampling. In: CVPR.
  pp. 5589--5598 (2020)

\bibitem{yang2018foldingnet}
Yang, Y., Feng, C., Shen, Y., Tian, D.: Foldingnet: Point cloud auto-encoder
  via deep grid deformation. In: CVPR. pp. 206--215 (2018)

\bibitem{yang2019learning}
Yang, Z., Wang, L.: Learning relationships for multi-view 3d object
  recognition. In: ICCV. pp. 7505--7514 (2019)

\bibitem{yi2016scalable}
Yi, L., Kim, V.G., Ceylan, D., Shen, I.C., Yan, M., Su, H., Lu, C., Huang, Q.,
  Sheffer, A., Guibas, L.: A scalable active framework for region annotation in
  3d shape collections. ACM Transactions on Graphics (ToG)  \textbf{35}(6),
  1--12 (2016)

\bibitem{yu2022point}
Yu, X., Tang, L., Rao, Y., Huang, T., Zhou, J., Lu, J.: Point-bert:
  Pre-training 3d point cloud transformers with masked point modeling. In:
  CVPR. pp. 19313--19322 (2022)

\bibitem{yu2024pedestrian}
Yu, Z., Li, L., Xie, J., Wang, C., Li, W., Ning, X.: Pedestrian 3d shape
  understanding for person re-identification via multi-view learning. IEEE
  Transactions on Circuits and Systems for Video Technology  (2024)

\bibitem{zha2023towards}
Zha, Y., Ji, H., Li, J., Li, R., Dai, T., Chen, B., Wang, Z., Xia, S.T.:
  Towards compact 3d representations via point feature enhancement masked
  autoencoders. arXiv preprint arXiv:2312.10726  (2023)

\bibitem{zhang2021pvt}
Zhang, C., Wan, H., Shen, X., Wu, Z.: Pvt: Point-voxel transformer for point
  cloud learning. arXiv preprint arXiv:2108.06076  (2021)

\bibitem{zhang2023deep}
Zhang, H., Wang, C., Tian, S., Lu, B., Zhang, L., Ning, X., Bai, X.: Deep
  learning-based 3d point cloud classification: A systematic survey and
  outlook. Displays  \textbf{79},  102456 (2023)

\bibitem{zhang2024pointgt}
Zhang, H., Wang, C., Yu, L., Tian, S., Ning, X., Rodrigues, J.: Pointgt: A
  method for point-cloud classification and segmentation based on local
  geometric transformation. IEEE Transactions on Multimedia  (2024)

\bibitem{zhang2023parameter}
Zhang, R., Wang, L., Wang, Y., Gao, P., Li, H., Shi, J.: Parameter is not all
  you need: Starting from non-parametric networks for 3d point cloud analysis.
  arXiv preprint arXiv:2303.08134  (2023)

\bibitem{zhao2021point}
Zhao, H., Jiang, L., Jia, J., Torr, P.H., Koltun, V.: Point transformer. In:
  ICCV. pp. 16259--16268 (2021)

\bibitem{zhou2021adaptive}
Zhou, H., Feng, Y., Fang, M., Wei, M., Qin, J., Lu, T.: Adaptive graph
  convolution for point cloud analysis. In: ICCV. pp. 4965--4974 (2021)

\bibitem{zhu2013facade}
Zhu, X.X., Shahzad, M.: Facade reconstruction using multiview spaceborne
  tomosar point clouds. IEEE Transactions on Geoscience and Remote Sensing
  \textbf{52}(6),  3541--3552 (2013)

\end{thebibliography}
	\end{document}